\documentclass[runningheads]{llncs}
\usepackage{graphicx}
\usepackage{amsmath}
\usepackage{amssymb}
\usepackage{multirow}
\usepackage{subfigure}

\newcommand{\bd}[1]{\textbf{#1}}
\newcommand{\app}{\raise.17ex\hbox{$\scriptstyle\sim$}}

\newcommand{\etal}[1]{{#1}~\emph{et~al.}}

\begin{document}
\title{Improve bone age assessment by learning from anatomical local regions}

\author{Dong Wang\inst{1} \and
Kexin Zhang\inst{2} \and
Jia Ding\inst{3} \and
Liwei Wang\inst{1,2}}

\authorrunning{Dong Wang et al.}
%
\institute{Center for Data Science, Peking University \and
Key Laboratory of Machine Perception, MOE, School of EECS, Peking University
\and
Yizhun Medical AI Co., Ltd
\\
\email{\{wangdongcis,zhangkexin,wanglw\}@pku.edu.cn}, 
\email{jia.ding@yizhun-ai.com}
}

\maketitle              

\begin{abstract}

Skeletal bone age assessment (BAA), as an essential imaging examination, aims at evaluating the biological and structural maturation of human bones. In the clinical practice, Tanner and Whitehouse (TW2) method is a widely-used method for radiologists to perform BAA. The TW2 method splits the hands into Region Of Interests (ROI) and analyzes each of the anatomical ROI separately to estimate the bone age. Because of considering the analysis of local information, the TW2 method shows accurate results in practice. Following the spirit of TW2, we propose a novel model called Anatomical Local-Aware Network (ALA-Net) for automatic bone age assessment. In ALA-Net, anatomical local extraction module is introduced to learn the hand structure and extract local information. Moreover, we design an anatomical patch training strategy to provide extra regularization during the training process. Our model can detect the anatomical ROIs and estimate bone age jointly in an end-to-end manner. The experimental results show that our ALA-Net achieves a new state-of-the-art single model performance of 3.91 mean absolute error (MAE) on the public available RSNA dataset. Since the design of our model is well consistent with the well recognized TW2 method, it is interpretable and reliable for clinical usage. 


\end{abstract}

\section{Introduction}

Skeletal bone age assessment (BAA) aims at evaluating the biological and structural maturation of human bones. 
As an essential imaging examination of clinical diagnosis, bone age assessment is acting in plenty of scenarios, such as identifying the growth disorder of the body or investigating endocrinology problems. 
In clinical practice, radiologists perform bone age assessment by analyzing the ossification patterns visually of the non-dominant hands in X-Ray images. The two most popular methods adopted by radiologists are Greulich and Pyle (G\&P) \cite{greulich1959radiographic} and Tanner and Whitehouse (TW2) \cite{tanner1975prediction}. 
The G\&P method assesses the entire hands as a whole, while the TW2 splits the hands into 20 Region Of Interests (ROI) and analyzes each of them separately to estimate the bone age. 
Since the TW2 method considers the analysis of local information, it shows more accurate results than the G\&P method~\cite{king1994reproducibility}. 
Nevertheless, these methods heavily rely on the expertise, and may be affected by subjective factors during observations and are considerably time-consuming, which are limitations for clinical applications.

Recently, accompanied by the development of deep learning and computer vision, the automatic BAA methods have shown great potential. Thanks to the public BAA dataset released by the Radiological Society of North America (RSNA) \cite{halabi2019rsna}, a series of deep learning methods are proposed and achieve impressive performance based on the dataset. 
Since the RSNA dataset only provides image-level annotations, most of these works adopt end-to-end training approaches and are not built to explicitly exploit the information of hand structure and local information. 
To make better use of anatomical local information, BoNet\cite{escobar2019hand} proposes to perform BAA by using hand keypoints heatmap as the input of network along with the hand image, to leverage the information of anatomical ROIs. 
The BoNet raises a useful viewpoint for using local information, while there are further limitations. The BoNet requires hand keypoints as input of the network during not only training but also inference, which should be provided by another network or human annotations in the paper. 

\begin{figure*}[t]
  \centering
  \subfigure[]{\includegraphics[width=1.3in]{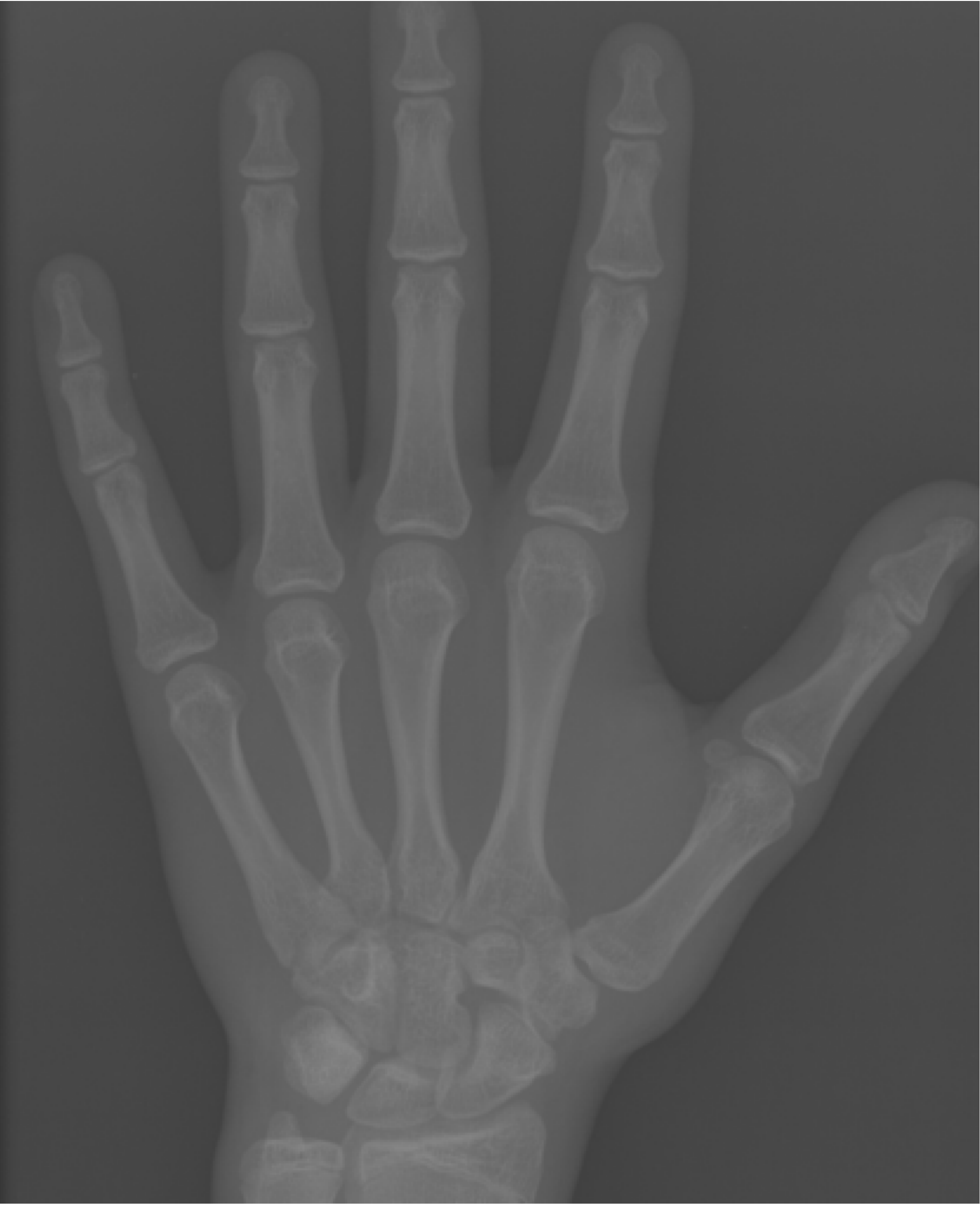}}
  \subfigure[]{\includegraphics[width=1.3in]{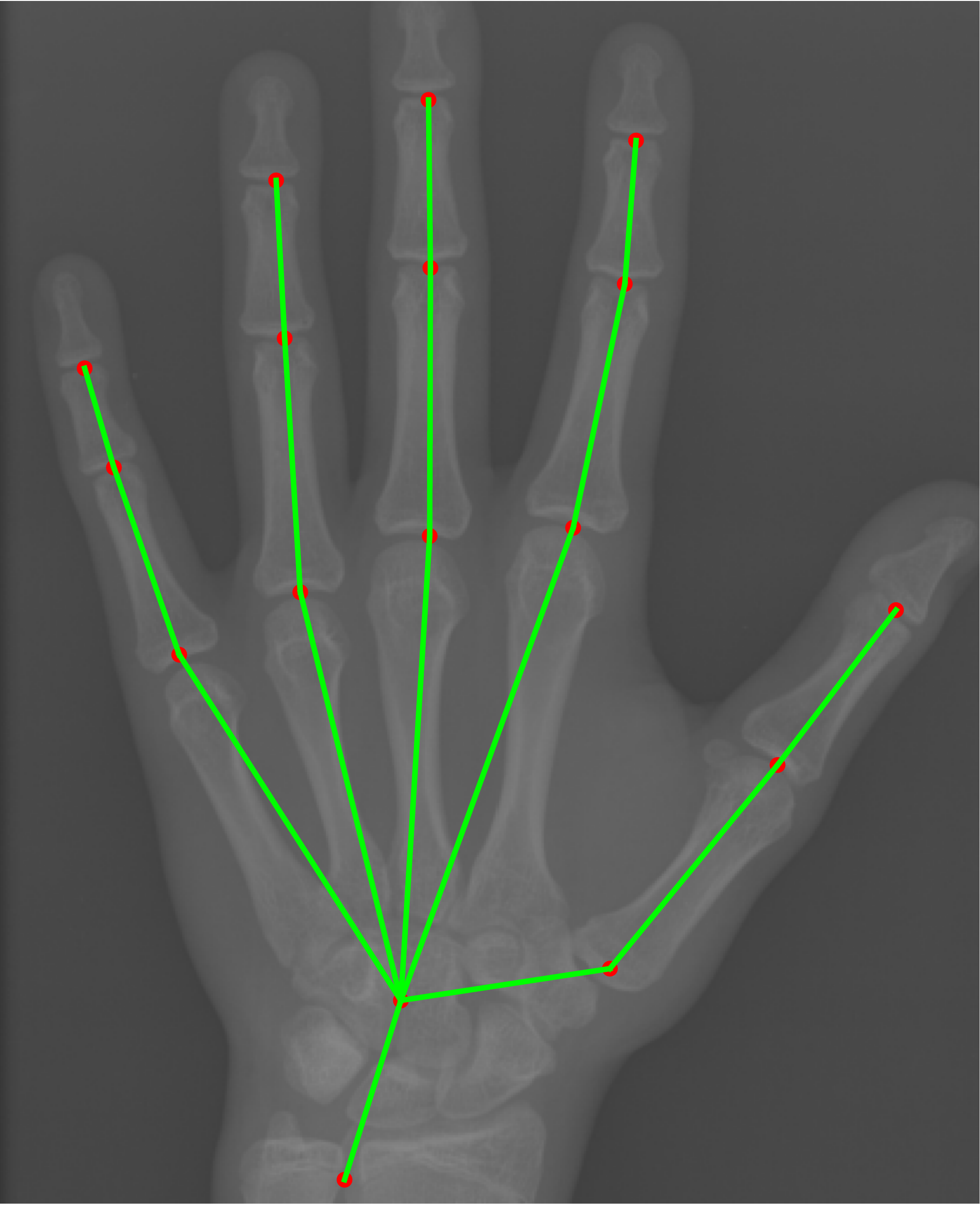}}
  \subfigure[]{\includegraphics[width=1.3in]{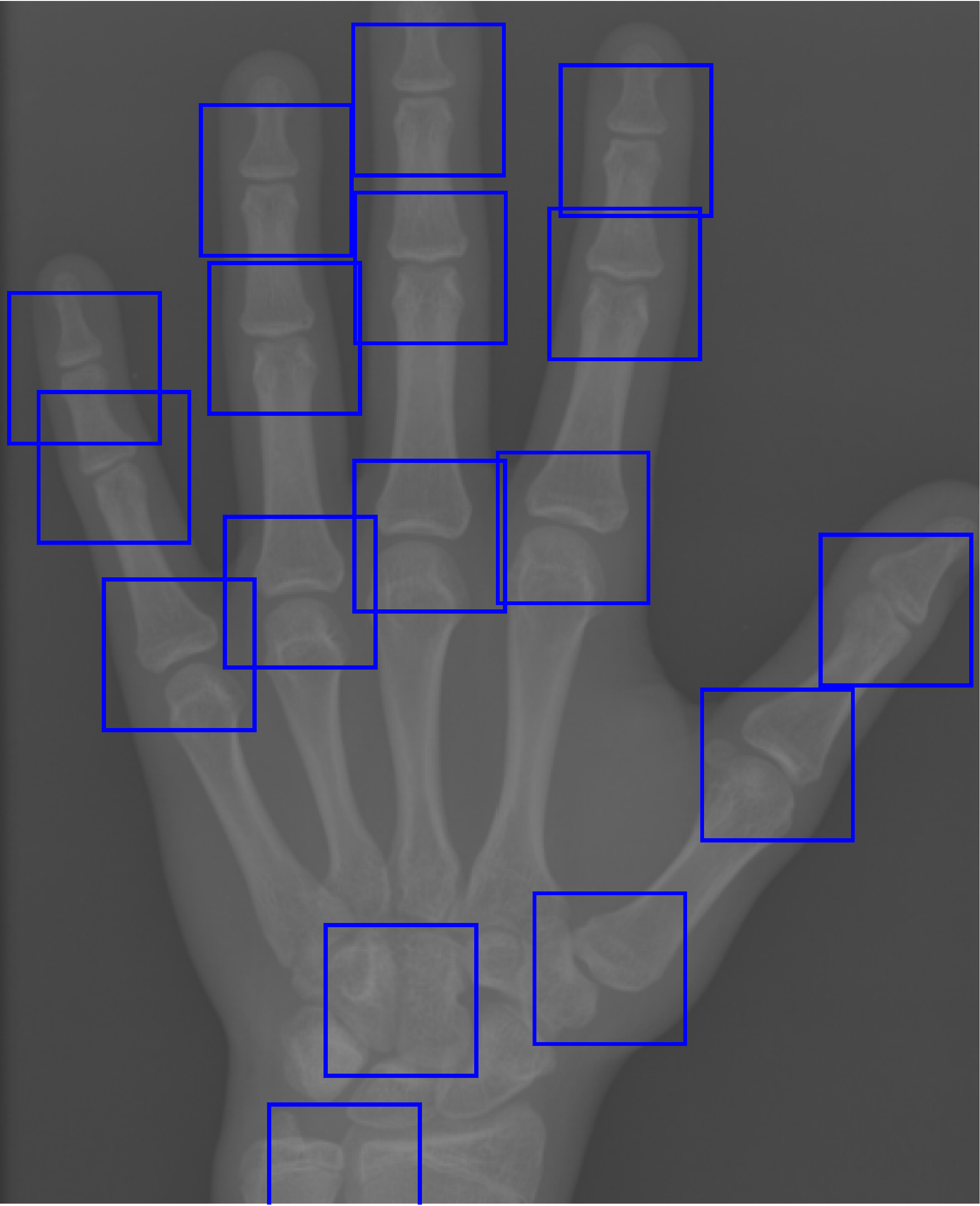}}
  \caption{ (a) Original image. (b) Hand keypoint annotation used in BoNet~\cite{escobar2019hand}. (c) We convert the keypoints to bounding boxes which are used in our method to learn hand structure and extract local information.}
  \label{fig1}
\end{figure*}

In this paper, to make better use of anatomical information and facilitate clinical usage, we propose a novel Anatomical Local-Aware Network (ALA-Net) for bone age assessment. We list the advantages of our method below:

\begin{enumerate}
    \item We propose an anatomical local extraction module to learn the hand structure and extract local information for BAA. Unlike BoNet, our ALA-Net can extract the anatomical ROIs (Fig.~\ref{fig1}) and evaluate the bone age jointly in an end-to-end manner. Benefiting from multi-task learning, the performances of anatomical region detection and BAA are significantly improved.
    \item Inspired by the TW2 method, we further propose an anatomical patch training strategy, which can be regarded as an effective regularization technique during the training process.
    \item Our approach is well consistent with the well recognized clinical assessment method (i.e., TW2) in BAA, which means the model structure is reliable and interpretable for clinical usage.

\end{enumerate}

Experiments show that ALA-Net outperforms other state-of-the-art single-model methods, and achieves 3.91 Mean Absolute Error (MAE) on the RSNA dataset. Meanwhile, the effectiveness of each component of the network design is validated through the ablation study.

\section{Related work}

Earlier deep learning based methods for bone age assessment adopt the end-to-end deep neural models, which take the whole hand image as input and make prediction for bone age. \etal{Larson}~\cite{larson2017performance} use a deep residual network with 50 layers to predict probability scores for each month. \etal{Spampinato}~\cite{SPAMPINATO201741} validate the effectiveness of deep CNNs pre-trained on general imagery in the bone age regression model. \etal{Torres}~\cite{torres2020empirical} introduce a carefully tuned architecture called GPNet for BAA. Even though achieving promising results, these models do not consider the local information of different bones. Accordingly, the models are lack of interpretability and have limited performance.

More recently, with the rapid progress of applying deep learning in medical image~\cite{esteva2017dermatologist,gulshan2016development,ding2017accurate,wang2020focalmix}, researchers have paid more attention to excavate the anatomical information for BAA or tried to incorporate human prior knowledge in the task. PRSNet~\cite{ji2019prsnet} uses a part selection module to select the most helpful hand parts for BAA and uses part relation module to model the multi-scale context information. \etal{Liu}~\cite{liu2019extract} propose to use the attention agent to discover the discriminative bone parts and extract features from these parts. Although the importance of bone part information is emphasized in their methods, there are also certain limitations. Due to the lack of prior knowledge of hand structure, the selected parts are more likely to the central regions of the hand, which can not capture the anatomical information of hand. To solve this problem, \etal{Escobar}~\cite{escobar2019hand} proposes BoNet, which uses hand pose estimation as the new task to extract local information. Specifically, BoNet uses the heatmap generated from hand keypoints as the input of the network along with the image. The manually annotated keypoints or pre-computed keypoints are required in both the training and inference stage, which leads to limited usability. Unlike BoNet, our ALA-Net learns the hand structure and anatomical ROIs itself, which boosts the performance through multi-task learning. Meanwhile, our end-to-end model is more convenient for practical usage since we do not need keypoints input during inference. Besides, there are also some works trying to estimate ages from MR images by deep learning methods~\cite{vstern2016automated,vstern2019automated}.


\section{Methodology}


\begin{figure}[t]
\includegraphics[width=\textwidth]{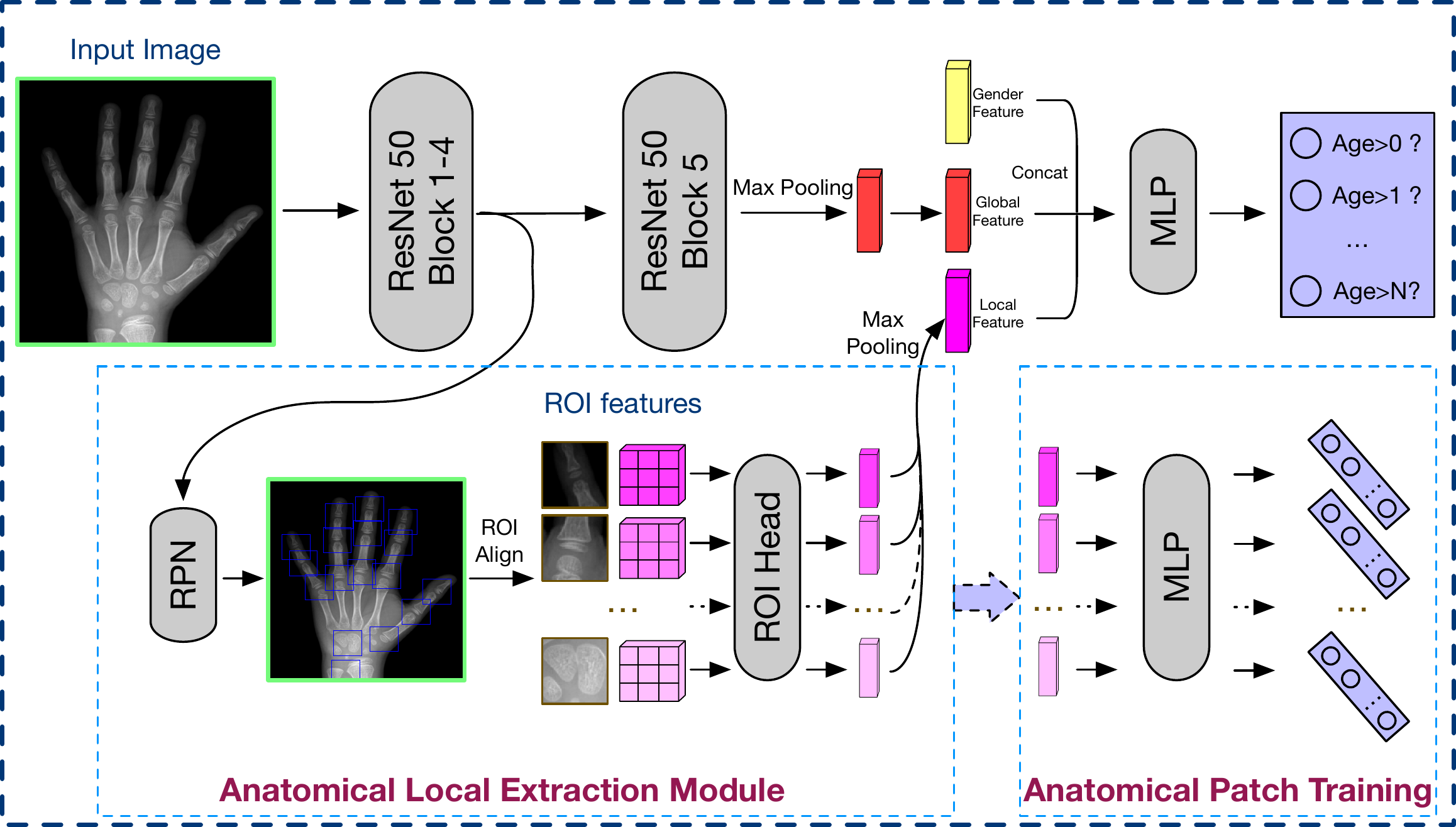}
\caption{Overview of our proposed anatomical local-aware network.}
\label{fig:net}
\end{figure}


\subsubsection{Overall Framework:}

The overview of our method is illustrated in Fig.~\ref{fig:net}. Taking the hand image as input, the network combines global feature, local feature, and gender feature to conduct bone age assessment. The global feature is generated by the ResNet-50, while the local feature is the output of our proposed local extraction module. We use the same gender feature as in \cite{ji2019prsnet}. We formulate the bone age assessment problem as an ordinal regression problem, and the final prediction is made by the multi-layer perceptron. Meanwhile, we use the anatomical patch training strategy to regularize the network during training. In the following sections, we describe the design of each part in detail.

\subsubsection{Anatomical Local Extraction Module:}

The anatomical structure provides rich information for the assessment of bone age. It has been proven that the local analysis of hand is essential in BAA in the TW2 method~\cite{king1994reproducibility}. Furthermore, from the perspective of multi-task training, the understanding of hand structure can be regarded as an additional task to promote the training of the network. Thus, we propose an anatomical local extraction module to perform local analysis and learn the hand structure at the same time.

As shown in Fig.~\ref{fig:net}, the anatomical local extraction module takes the output feature of the 4th block of ResNet as input, and a region proposal network (RPN) is applied to generate ROI proposals. We only leave the top 17 ROIs according to the definition of hand keypoints. Then, the ROI align module extracts local features for the ROIs. We use a shared ROI head with four convolution layers and one fully-connection layer to transform the ROI features. Finally, we apply max pooling across the ROI features to aggregate the transformed ROI features and output the local feature for further analysis of BAA.

The original annotations of hand pose are a series of points, and we need to use boxes for local extraction. Hence, during the training time, we replace the point annotations by boxes centered at the points (Fig.~\ref{fig1}). We use the same loss function as in \cite{ren2015faster} to train RPN, and we denote the loss function of RPN as $L_{RPN}$.

\subsubsection{Ordinal Regression:} 

The commonly used $L_2$ loss varies widely in scale during the training process, which causes difficulty for convergence in multi-task training. To tackle this problem, we use ordinal regression~\cite{niu2016ordinal} to predict bone age. In the ordinal regression method, the regression problem of $K$ ranks is transformed into a $K-1$ binary classification problem. For each rank $r \in \{ 0, 1,..., K - 2 \}$ and an example with rank $y$, a binary classifier is adopted to classify whether the $y$ is larger than $r$. We implement the prediction by using $K-1$ dimension logits followed by the sigmoid activation function, and it is intuitive to employ the binary cross-entropy loss as loss function,

\begin{equation}
    L_{ord}(o, y) = \frac{1}{K-1}\sum_{k=0}^{K-2} BCE(o_k, 1\{y>k\}),
\end{equation}
where $o$ is the $K-2$ dimension output probabilities, $y$ is the target rank, and $BCE$ denotes binary cross entropy loss. During the inference time, we could simply add the probabilities of all ranks to output the predicted bone age.

\subsubsection{Anatomical Patch Training:} 

Crop augmentation is a frequently-used technique for training neural networks. The augmentation method can enrich the dataset and make the network more robust to the small variations of the input images. For general images, it is a common practice to crop the images in a random region. While in our task, to integrate anatomical knowledge into our network, we propose to train BAA on the detected anatomical ROIs as a data augmentation method. As shown in the bottom right part in Fig.~\ref{fig:net}, during the training stage, we use a multi-layer perceptron to predict bone ages on the features of each ROI patch and train the model by ordinal regression losses described above. The 17 ROIs detected by the local extraction module are used in this step.

The anatomical patch training method is inspired by the clinical practice of the TW2 method. The TW2 method performs intensive local analysis on different bone parts and combine the results to make a final decision. Following this spirit, our anatomical patch training strategy aims at teaching the network to evaluate the growth of each bone, which promotes the performance of BAA. In consequence, the anatomical patch training method is well consistent with the motivation of the TW2 method.

\subsubsection{Loss function:}

Our network can be trained end-to-end using a compound loss function:

\begin{equation}
    L_{total} = L_{ord} + L_{ord}^{patch} + L_{RPN},
\end{equation}
where $L_{ord}$ and $L_{RPN}$ are introduced in previous subsections, and $L_{ord}^{patch}$ is the ordinal regression loss for anatomical patch training.

\section{Experiments}

\subsection{Experimental Setup}

\subsubsection{Dataset and Metric:} 

We conduct all experiments on the RSNA Pediatric Bone Age Challenge dataset\cite{halabi2019rsna}. The RSNA dataset consists of 12611 images for training, 1425 images for validation, and 200 images for testing. We use the hand keypoints annotations provided by \cite{escobar2019hand}. We report all the results on the test set. The Mean Absolute Error (MAE) between ground-truth age and predicted age is reported to evaluate the model performance, and the unit of measurement is month.

\subsubsection{Network:} 

 We implement the network using PyTorch. We resize the long side of the input images to 512 pixels and keep the original aspect ratios. The box size of anatomical ROI is set to be $64\times64$. During the training stage, we use Adam optimizer to optimize the network. The learning rate is set to be $0.001$ initially and reduced by a factor of 10 at 30k and 40k iteration, respectively. We train the network for 50k iterations and use a batch size of 32. Apart from the anatomical patch training method, we use horizontal flipping and random scaling as data augmentation. The total training process costs about 5 hours with 2 TITAN RTX GPU cards, and the inference time is 0.036s per image.
 

\begin{table}[]
\setlength{\tabcolsep}{2mm}
\renewcommand\arraystretch{1.2}
\centering
\caption{Ablation study of bone age assessment.}
\label{tab:ablation_baa}
\begin{tabular}{c|c|c||l}
\hline
ResNet-50    & Local extraction    & Patch training        & MAE         \\
\hline
\checkmark    &                    &                    & 4.92        \\
\checkmark    & \checkmark        &                    & 4.53        \\
\checkmark    &                    & \checkmark            & 4.49        \\
\checkmark    & \checkmark        & \checkmark            & \bd{3.91}    \\
\hline
\end{tabular}
\end{table}

 
\subsection{Ablation Study}

\subsubsection{Bone Age Assessment:}    

We first study the effect of each component in the ALA-Net, i.e., anatomical local extraction module and anatomical patch training. The experimental results are shown in Table~\ref{tab:ablation_baa}. Without both of these components, the ALA-Net degrades to the ResNet-50, which produces the MAE score of 4.92. When the anatomical local extraction module is applied to this model, the MAE score becomes 4.53, which is an improvement of 0.39 in MAE. The improvement confirms the necessity of learning anatomical structures and extracting local features. We also test the performance of ResNet-50 with the anatomical patch training strategy alone. In this circumstance, the network can not generate ROI proposals itself, hence we use the annotated ROIs during the training stage to replace the proposals predicted by the region proposal network. The anatomical patch training strategy performs a strong regularization for the network, and improves the MAE score to 4.49. Combining these two components, our ALA-Net achieves 3.91 MAE.


\begin{table}[]
\setlength{\tabcolsep}{3mm}
\renewcommand\arraystretch{1.2}
\centering
\caption{Ablation study of ROI detection.}
\label{tab:ablation_roi}
\begin{tabular}{c|c|c|c}
\hline
Method        & AP(\%)& AP50(\%)    & AP75(\%)     \\
\hline
RPN            & 86.2    & 97.9        & 96.2        \\
ALA-Net(our)& 89.2    & 98.0        & 98.0        \\

\hline
\end{tabular}
\end{table}


\subsubsection{Anatomical ROI Detection:}

Our model detects anatomical ROIs and performs bone age assessment jointly. We observe that the ROI detection also benefits from the multi-task learning experimentally. We use mean Average Precision (AP) at Intersection over Union (IoU) at $[0.5 : 0.05: 0.95]$ to measure the performance. The results are shown in Table~\ref{tab:ablation_roi}. We use a vanilla RPN as the baseline, and our model outperforms the RPN significantly, which demonstrates that our model has a better understanding of the hand structure.

\begin{table}
\setlength{\tabcolsep}{2mm}
\renewcommand\arraystretch{1.2}
\centering
\caption{Comparison with state-of-the-art methods on the RSNA dataset. Notations: ``Global'' -- using global information; ``Local'' -- using local information.}
\label{tab:sota}
\begin{tabular}{l|c|c||c}
\hline
Method    & Global    & Local & MAE \\
\hline
Human\cite{larson2017performance}&&&7.32\\
Larson\cite{larson2017performance}&\checkmark&&6.24\\
Ren\cite{ren2019regression}&\checkmark&&5.20\\
Iglovikov\cite{iglovikov2018paediatric}&\checkmark&\checkmark&4.97\\
PRSNet\cite{ji2019prsnet}&\checkmark&\checkmark&4.49\\
AR-CNN\cite{liu2019extract}&\checkmark&\checkmark&4.38\\
BoNet\cite{escobar2019hand}&\checkmark&\checkmark&4.14\\
ALA-Net(Our)&\checkmark&\checkmark&\bd{3.91}\\
\hline
\end{tabular}
\end{table}


\subsection{Comparison with State-of-the-Arts}

In this subsection, we compare our network with previous state-of-the-art single-model methods. The results are shown in Table~\ref{tab:sota}. For each method, we show whether it uses global or local information for BAA. The human performance of bone age assessment is 7.32 MAE~\cite{larson2017performance}. The approaches only using global feature have a slightly better performance than humans. Our ALA-Net yields better performance since we consider both global and local features. Compared with the approaches using both global and local features, our model still has significant advantages. It is attributed to not only the learning of the anatomical structure of hand but also the effectiveness of anatomical local information. As a result, our method achieves a performance gain of 0.23 MAE over the previous SOTA~\cite{escobar2019hand}, which is a relative improvement of $5.6\%$.

\begin{figure}[!t]
\includegraphics[width=\textwidth]{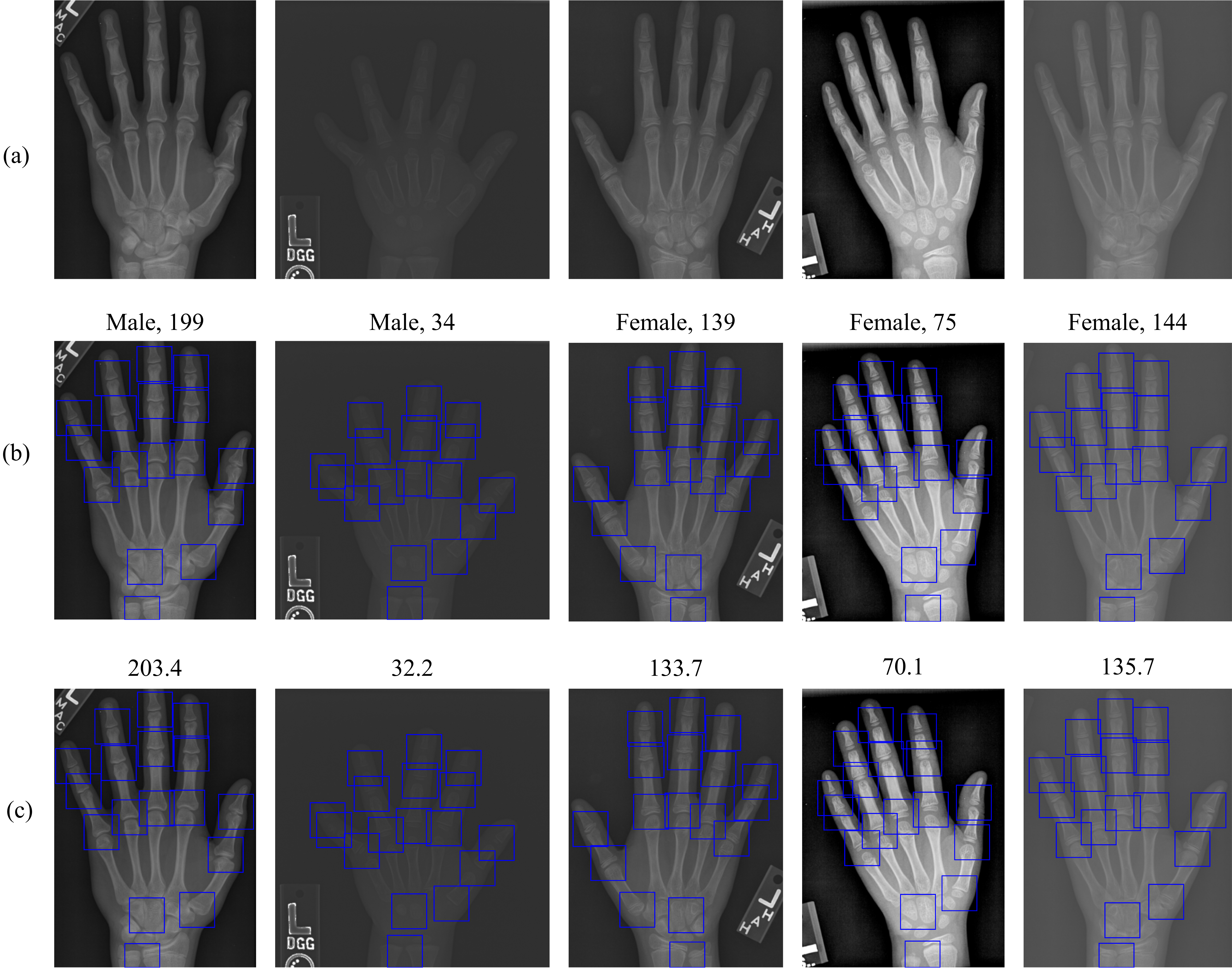}
\caption{Case study. (a) The original image. (b) Gender and ground truth age, ROIs. (c) Predicted age and ROIs. }
\label{fig:vis}
\end{figure}

For a better understanding of our method, we show some randomly sampled examples in Fig.~\ref{fig:vis}. We can observe that the detected anatomical ROIs are close to the annotations, which leads to accurate BAA results.


\section{Conclusion}

In this work, we propose a novel model called ALA-Net for bone age assessment. The anatomical local extraction model is introduced to learn the structure of hands and extract local information. The anatomical patch training strategy further provides a regularization term to enhance the performance of bone age assessment. Experimental results on the RSNA dataset demonstrates that our model achieves a new state-of-the-art result. Moreover, since the design of our model takes human priors into account, it is well interpretable and reliable for medical practice. 
Our work emphasizes the importance of medical prior knowledge for model design. We encourage further exploration of how to integrate medical knowledge into medical image analysis.

%
%

\bibliographystyle{splncs04}
\bibliography{mybibliography}

\begin{thebibliography}{10}
\providecommand{\url}[1]{\texttt{#1}}
\providecommand{\urlprefix}{URL }
\providecommand{\doi}[1]{https://doi.org/#1}

\bibitem{ding2017accurate}
Ding, J., Li, A., Hu, Z., Wang, L.: Accurate pulmonary nodule detection in
  computed tomography images using deep convolutional neural networks. In:
  International Conference on Medical Image Computing and Computer-Assisted
  Intervention. pp. 559--567. Springer (2017)

\bibitem{escobar2019hand}
Escobar, M., Gonz{\'a}lez, C., Torres, F., Daza, L., Triana, G., Arbel{\'a}ez,
  P.: Hand pose estimation for pediatric bone age assessment. In: International
  Conference on Medical Image Computing and Computer-Assisted Intervention. pp.
  531--539. Springer (2019)

\bibitem{esteva2017dermatologist}
Esteva, A., Kuprel, B., Novoa, R.A., Ko, J., Swetter, S.M., Blau, H.M., Thrun,
  S.: Dermatologist-level classification of skin cancer with deep neural
  networks. Nature  \textbf{542}(7639), ~115 (2017)

\bibitem{greulich1959radiographic}
Greulich, W.W., Pyle, S.I.: Radiographic atlas of skeletal development of the
  hand and wrist. Stanford University Press (1959)

\bibitem{gulshan2016development}
Gulshan, V., Peng, L., Coram, M., Stumpe, M.C., Wu, D., Narayanaswamy, A.,
  Venugopalan, S., Widner, K., Madams, T., Cuadros, J., et~al.: Development and
  validation of a deep learning algorithm for detection of diabetic retinopathy
  in retinal fundus photographs. Jama  \textbf{316}(22),  2402--2410 (2016)

\bibitem{halabi2019rsna}
Halabi, S.S., Prevedello, L.M., Kalpathy-Cramer, J., Mamonov, A.B., Bilbily,
  A., Cicero, M., Pan, I., Pereira, L.A., Sousa, R.T., Abdala, N., et~al.: The
  rsna pediatric bone age machine learning challenge. Radiology
  \textbf{290}(2),  498--503 (2019)

\bibitem{iglovikov2018paediatric}
Iglovikov, V.I., Rakhlin, A., Kalinin, A.A., Shvets, A.A.: Paediatric bone age
  assessment using deep convolutional neural networks. In: Deep Learning in
  Medical Image Analysis and Multimodal Learning for Clinical Decision Support,
  pp. 300--308. Springer (2018)

\bibitem{ji2019prsnet}
Ji, Y., Chen, H., Lin, D., Wu, X., Lin, D.: {PRSNet}: Part relation and
  selection network for bone age assessment. In: International Conference on
  Medical Image Computing and Computer-Assisted Intervention. pp. 413--421.
  Springer (2019)

\bibitem{king1994reproducibility}
King, D., Steventon, D., O'sullivan, M., Cook, A., Hornsby, V., Jefferson, I.,
  King, P.: Reproducibility of bone ages when performed by radiology
  registrars: an audit of tanner and whitehouse ii versus greulich and pyle
  methods. The British journal of radiology  \textbf{67}(801),  848--851 (1994)

\bibitem{larson2017performance}
Larson, D.B., Chen, M.C., Lungren, M.P., Halabi, S., Stence, N.V., Langlotz,
  C.P.: Performance of a deep-learning neural network model in assessing
  skeletal maturity on pediatric hand radiographs. Radiology  \textbf{287}(1),
  313--322 (2017)

\bibitem{liu2019extract}
Liu, C., Xie, H., Liu, Y., Zha, Z., Lin, F., Zhang, Y.: Extract bone parts
  without human prior: end-to-end convolutional neural network for pediatric
  bone age assessment. In: International Conference on Medical Image Computing
  and Computer-Assisted Intervention. pp. 667--675. Springer (2019)

\bibitem{niu2016ordinal}
Niu, Z., Zhou, M., Wang, L., Gao, X., Hua, G.: Ordinal regression with multiple
  output cnn for age estimation. In: Proceedings of the IEEE conference on
  computer vision and pattern recognition. pp. 4920--4928 (2016)

\bibitem{ren2015faster}
Ren, S., He, K., Girshick, R., Sun, J.: Faster r-cnn: Towards real-time object
  detection with region proposal networks. In: Advances in neural information
  processing systems. pp. 91--99 (2015)

\bibitem{ren2019regression}
Ren, X., Li, T., Yang, X., Wang, S., Ahmad, S., Xiang, L., Stone, S.R., Li, L.,
  Zhan, Y., Shen, D., et~al.: Regression convolutional neural network for
  automated pediatric bone age assessment from hand radiograph. IEEE Journal of
  Biomedical and Health Informatics  \textbf{23}(5),  2030--2038 (2019)

\bibitem{SPAMPINATO201741}
Spampinato, C., Palazzo, S., Giordano, D., Aldinucci, M., Leonardi, R.: Deep
  learning for automated skeletal bone age assessment in x-ray images. Medical
  Image Analysis  \textbf{36},  41 -- 51 (2017).
  \doi{https://doi.org/10.1016/j.media.2016.10.010}

\bibitem{vstern2016automated}
{\v{S}}tern, D., Payer, C., Lepetit, V., Urschler, M.: Automated age estimation
  from hand mri volumes using deep learning. In: International Conference on
  Medical Image Computing and Computer-Assisted Intervention. pp. 194--202.
  Springer (2016)

\bibitem{vstern2019automated}
{\v{S}}tern, D., Payer, C., Urschler, M.: Automated age estimation from mri
  volumes of the hand. Medical image analysis  \textbf{58},  101538 (2019)

\bibitem{tanner1975prediction}
Tanner, J., Whitehouse, R., Marshall, W., Carter, B.: Prediction of adult
  height from height, bone age, and occurrence of menarche, at ages 4 to 16
  with allowance for midparent height. Archives of disease in childhood
  \textbf{50}(1),  14--26 (1975)

\bibitem{torres2020empirical}
Torres, F., Gonz{\'a}lez, C., Escobar, M.C., Daza, L., Triana, G.,
  Arbel{\'a}ez, P.: An empirical study on global bone age assessment. In: 15th
  International Symposium on Medical Information Processing and Analysis. vol.
  11330, p. 113300E. International Society for Optics and Photonics (2020)

\bibitem{wang2020focalmix}
Wang, D., Zhang, Y., Zhang, K., Wang, L.: Focalmix: Semi-supervised learning
  for 3d medical image detection. arXiv preprint arXiv:2003.09108  (2020)

\end{thebibliography}

\end{document}